\documentclass[10pt,twocolumn,letterpaper]{article}

\usepackage{titlesec}
\usepackage{cvpr}
\usepackage{times}
\usepackage{epsfig}
\usepackage{graphicx}
\usepackage{amsmath}
\usepackage{amssymb}
\usepackage{graphicx}
\usepackage{epstopdf}
\usepackage{subfig}
\usepackage{makecell}
\usepackage{multirow}
\usepackage{array}
\usepackage{bm}
\usepackage{color}
\usepackage{caption}
\newcommand{\PreserveBackslash}[1]{\let\temp=\\#1\let\\=\temp}
\newcolumntype{C}[1]{>{\PreserveBackslash\centering}p{#1}}
\newcolumntype{R}[1]{>{\PreserveBackslash\raggedleft}p{#1}}
\newcolumntype{L}[1]{>{\PreserveBackslash\raggedright}p{#1}}

\usepackage[pagebackref=true,breaklinks=true,letterpaper=true,colorlinks,bookmarks=false]{hyperref}

\cvprfinalcopy 
\def\cvprPaperID{1264} 

\ifcvprfinal\fi
\begin{document}

\title{End-to-end Flow Correlation Tracking with Spatial-temporal Attention}

\author{Zheng Zhu$^{1,2,3}$, Wei Wu$^{3}$, Wei Zou$^{1,2}$, Junjie Yan$^{3}$\\
$^{1}$Institute of Automation, Chinese Academy of Sciences, Beijing,
People{\textquoteright}s Republic of China\\
$^{2}$University of Chinese Academy of Sciences, Beijing, People{\textquoteright}s
Republic of China\\
$^{3}$SenseTime Group Limited, Beijing, People{\textquoteright}s
Republic of China}

\maketitle

\begin{abstract}
Discriminative correlation filters (DCF) with deep convolutional features have achieved favorable performance in recent tracking benchmarks. However, most of existing DCF trackers only consider appearance features of current frame, and hardly benefit from motion and inter-frame information. The lack of temporal information degrades the tracking performance during challenges such as partial occlusion and deformation.
In this work, we focus on making use of the rich flow information in consecutive frames to improve the feature representation and the tracking accuracy.  Firstly,  individual components, including optical flow estimation, feature extraction, aggregation and correlation filter tracking are formulated  as special layers in network. To the best of our knowledge, this is the first work to jointly train flow and tracking task in a deep learning framework. Then the historical feature maps at predefined intervals are warped and aggregated with current ones by the guiding of flow. For adaptive aggregation, we propose a novel spatial-temporal attention mechanism.
 Extensive experiments are performed on four challenging tracking datasets: OTB2013, OTB2015, VOT2015 and VOT2016, and the proposed method achieves superior results on these benchmarks.

\end{abstract}

\section{Introduction}

Visual object tracking, which tracks a specified target in a changing video sequence automatically, is a fundamental problem in many topics such as visual analysis \cite{c5}, automatic driving \cite{c6}, pose estimation \cite{c8} and \etal. A core problem of tracking is how to detect and locate the object accurately in changing scenarios with occlusions, shape deformation, illumination variations and \etal \cite{c9,c10}.

\begin{figure}[!tp]
\setlength{\abovecaptionskip}{0.5cm}
\setlength{\belowcaptionskip}{0.4cm}
  \centering
  \includegraphics[width=0.9\linewidth]{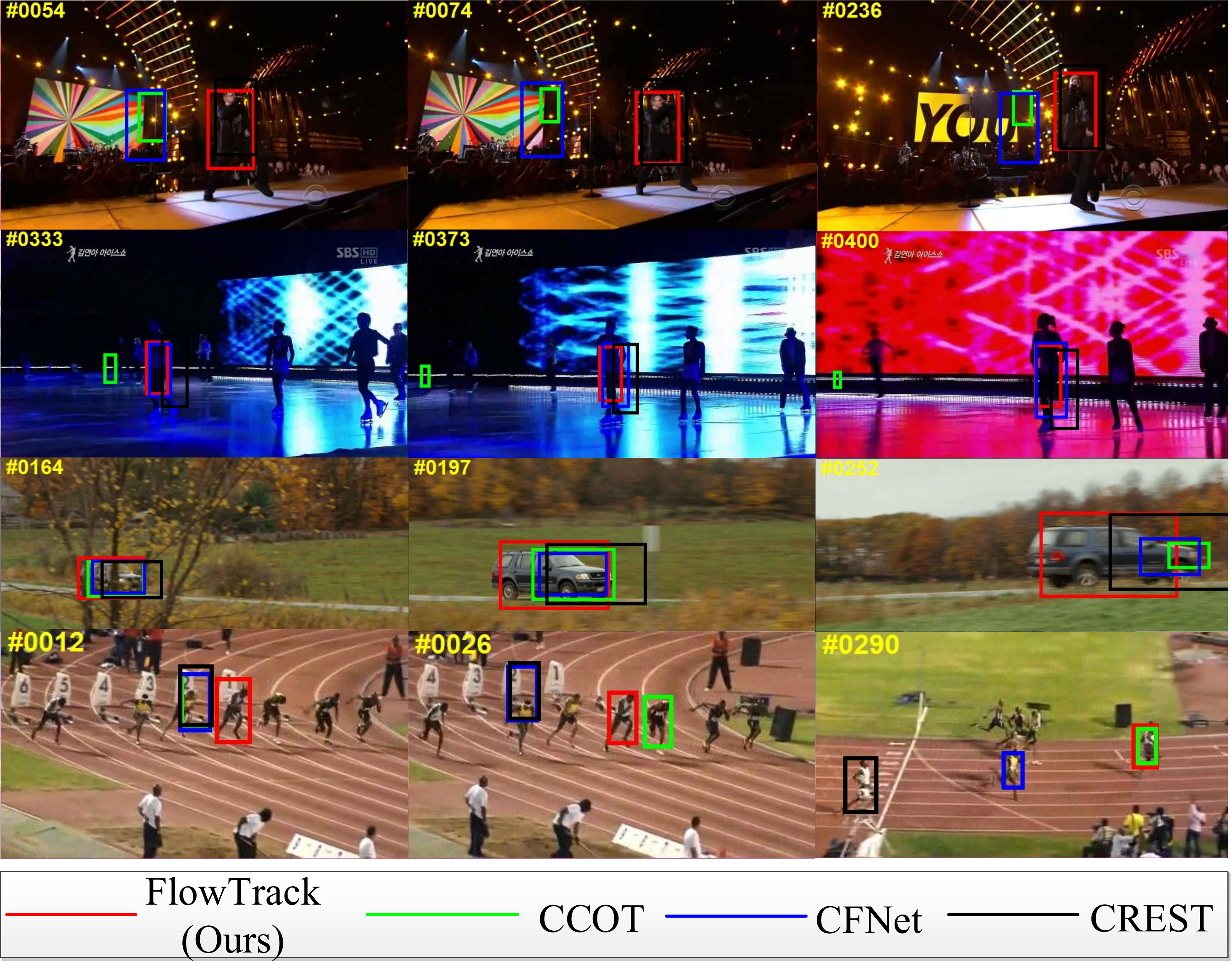}
  \caption{Tracking results comparison of our approach with three state-of-the-art trackers in the challenging scenarios. Best viewed on color display.}
  \label{tracking_results}
\end{figure}

 Recently, significant attention has been paid to discriminative correlation filters (DCF) based methods for visual tracking such as KCF\cite{c15,c25}, SAMF\cite{c35}, LCT \cite{c17}, MUSTer \cite{muster}, SRDCF \cite{c16} and CACF \cite{cacf}. Most of these methods use handcrafted features, which hinder their accuracy and robustness.  Inspired by the success of CNN in object recognition \cite{c18, c19, c20}, the visual tracking community has been focus on the deep trackers that exploit the strength of CNN in recent years. Representative deep trackers include DeepSRDCF \cite{c32}, HCF \cite{c27}, SiamFC \cite{c33} and CFNet \cite{cfnet}. However, most existing trackers only consider appearance features of current frame, and can hardly benefit from motion and inter-frame information. The lack of temporal information degrades the tracking performance during challenges such as partial occlusion and deformation. Although some trackers utilize optical flow to upgrade performance\cite{sint, best paper}, the flow feature is off-the-shelf and not trained end-to-end. These methods do not take full advantage of flow information, so achieved results may be suboptimal.


In this paper, we develop an end-to-end flow correlation tracking framework (FlowTrack) to utilize both the flow information and appearance features. Specifically, we formulate the optical flow estimation, feature extraction, aggregation and correlation filter tracking as special layers in network, which enables end-to-end learning. Then the previous frames are warped to specified frame by the guiding of flow information, and they are aggregated for consequent correlation filter tracking. For adaptive aggregation, a novel spatial-temporal attention mechanism is developed.

Features from different frames provide diverse information for same object instance, such as different viewpoints, deformation and varied illuminations. So appearance feature for tracked object can be enhanced by aggregating these feature maps. Figure~\ref{tracking_results} shows four challenging benchmark sequences which undergo illumination variation, viewpoint changes and deformation. The FlowTrack can handle these challenges due to the aggregation of diverse feature maps. Note that the features of the same object instance are usually not spatially aligned across frames due to video motion. A naive feature fusion may even deteriorate the performance. This suggests that it is critical to model the motion during learning. In the FlowTrack, the flow estimation and feature aggregation are end-to-end trained using large-scale tracking dataset.

We validate the effectiveness of our approach on four object tracking benchmarks: OTB2013\cite{c14}, OTB2015\cite{c9}, VOT2015\cite{c10} and VOT2016\cite{c11}. On the challenging OTB2013 and OTB2015, our object tracking framework obtain 0.689 and 0.655 in area under curve (AUC), respectively. On the VOT2015 and VOT2016, the FlowTrack both ranks \emph{1st} in expected average overlap (EAO) according to the latest VOT rules, while performing at 12FPS.
\paragraph{Contributions}
The contributions of this paper can be summarized in three folds as follows:

1, We develop an end-to-end flow correlation tracking framework  to improve the feature representation and the tracking accuracy. To the best of our knowledge, this is the first work to jointly train flow and tracking task in a deep learning framework.

2, A novel spatial-temporal attention mechanism is proposed, which can adaptively aggregate the warped and current feature maps.

3, In experiments of OTB2013, OTB2015, VOT2015 and VOT2016, the proposed tracking algorithm performs favorably against existing state-of-the-art methods in terms of accuracy and robustness.

\section{Related works}
Visual tracking is a significant problem in computer vision systems and a series of approaches have been proposed in recent years. Since our main contribution is an end-to-end framework for flow correlation tracking, we give a brief review on three directions closely related to this work: DCF-based trackers, CNN-based trackers, and optical flow in visual recognition.

\subsection{DCF trackers}
In recent tracking community, significant attention has been paid to discriminative correlation filters (DCF) based methods \cite{c15, c16, c17, c35,mosse,c25,c45,mcpf,cacf,bacf,muster,sct,CSR-DCF} because of their efficiency and expansibility. MOSSE \cite{mosse}, CSK \cite{c25} and KCF \cite{c15} are conventional DCF trackers. Many improvements for DCF tracking approaches have been proposed, such as SAMF \cite{c35} and fDSST\cite{c45} for scale changes, CN \cite{cn} and Staple \cite{c26} taking color information into account, LCT \cite{c17} and MUSTer \cite{muster} for long-term tracking, SRDCF \cite{c16} and CACF \cite{cacf}to mitigate boundary effects. The better performance is obtained, the more time DCF based tracker costs.  Most of these methods use handcrafted features, which hinder their accuracy and robustness.
Inspired by the success of convolution neural networks (CNN) in object classification \cite{c18, c19}, detection \cite{c20} and segmentation tasks\cite{c21}, researchers in tracking community have started to focus on the deep trackers that exploit the strength of CNN. Since DCF provides an excellent framework for recent tracking research, the popular trend is the combination of DCF framework and CNN features. In HCF \cite{c27} and HDT \cite{c28}, CNN are employed to extract features instead of handcrafted features, and final tracking results are obtained by combining hierarchical response and hedging weak trackers, respectively. DeepSRDCF \cite{c32} exploits shallow CNN features in a spatially regularized DCF framework. In above mentioned methods, the chosen CNN features are always pre-trained in different tasks and individual components in tracking systems are learned separately. So the achieved tracking results may be suboptimal. It is worth noting that CFNet \cite{cfnet} and DCFNet \cite{dcfnet} interpret the correlation filters as a differentiable layer in a Siamese tracking framework, thus achieving an end-to-end representation learning. The main drawback is its unsatisfying performance.

\subsection{CNN-based trackers}

Except for the combination of DCF framework and CNN features, another trend in deep trackers is to design the tracking networks and pre-train them in order to learn the target-specific features and handle the challenges for each new video. Bertinetto et.al \cite{c33} propose a fully convolutional Siamese network (SiamFC) to estimate the feature similarity region-wise between two frames. The network is trained off-line and evaluated without any online fine-tuning. Similar to SiamFC, in GOTURN tracker \cite{c34}, the motion between successive frames is predicted using a deep regression network. MDNet \cite{c29} trains a small-scale network by multi-domain methods, thus separating domain independent information from domain-specific layers. C-COT \cite{c30} and ECO \cite{eco} employ the implicit interpolation method to solve the learning problem in the continuous spatial domain, where ECO is an improved version of C-COT in performance and speed. CREST \cite{crest} treats tracking process as convolution and applies residual learning to take appearance changes into account. Similarly, UCT \cite{uct} treats feature extractor and tracking process both as convolution operation and trains them jointly, enabling learned CNN features tightly coupled to tracking process. All these trackers only consider appearance features in current frame and can hardly benefit from motion and inter-frame information.
In this paper, we make full use of these information by aggregating flow and Siamese tracking in an end-to-end framework.
\subsection{Optical flow for visual recognition}

Flow information has been exploited to be helpful in computer vision tasks. In pose estimation \cite{flow-pose}, optical flow is used to align heatmap predictions from neighbouring frames. \cite{flow-predict} applies flow to the current frame to predict next frame. In \cite{flow-transfer}, flow is used to explicitly model how image attributes vary with its deformation.
DFF \cite{dff} and FGFA \cite{fgfa} utilize flow information to speed up vision recognition (segmentation and video detection) and upgrade performance, respectively. In DFF, expensive convolutional sub-network is performed only on sparse key frames, and their deep feature maps are propagated to other frames via a flow field. In FGFA, nearby features are aggregated along the motion paths using flow information, thus improving the video recognition accuracy. Recently, some trackers also utilize optical flow to upgrade performance\cite{sint, best paper}, while the flow feature is off-the-shelf and not trained end-to-end. Since the features of the same object instance are usually not spatially aligned across frames due to video motion, a naive feature fusion may not gain performance.

\section{End-to-end flow correlation tracking}
In this section, flow correlation network is given at first to describe the overall training architecture. Then we introduce the correlation filter layer and the aggregation of optical flow. In order to adaptively weight the aggregated frames at each spatial location and temporal channels, a novel spatial-temporal attention mechanism is designed. At last, online tracking is described consisting of model updating and scales.

\subsection{Training network architecture}
\begin{figure*}[!tp]
  \centering
  \includegraphics[width=0.8\linewidth]{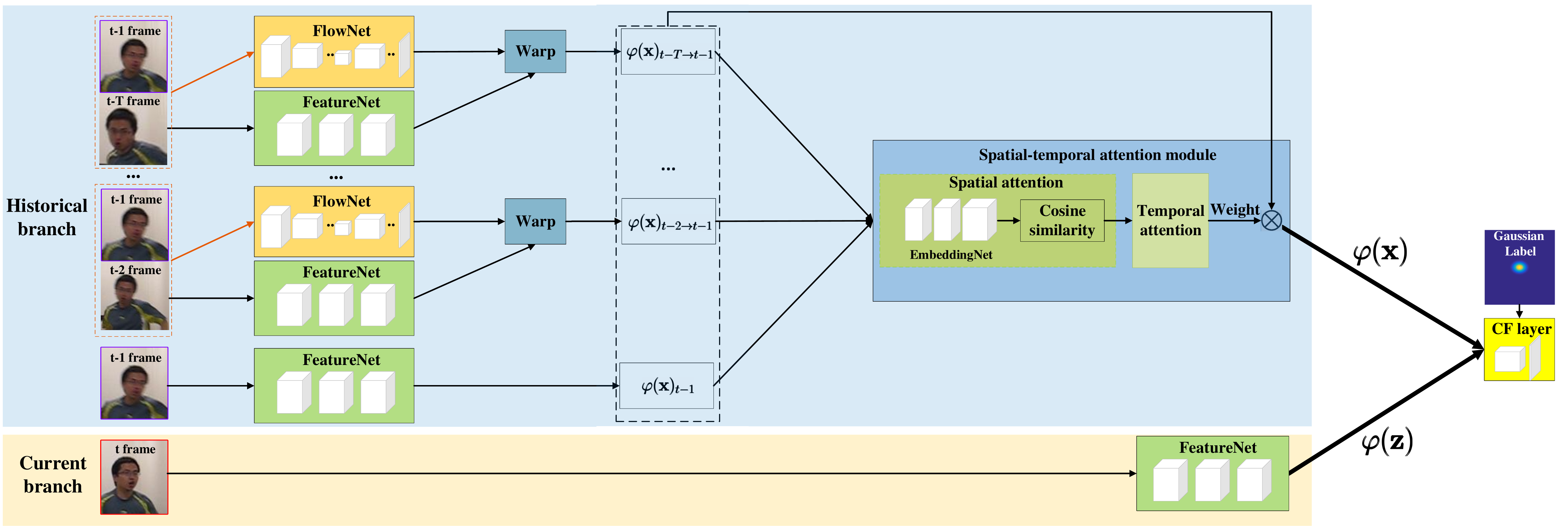}
  \caption{The overall training network. The network adopts Siamese architecture consisting of historical and current branches. The dashed boxes in left part represent concatenating two input frames for FlowNet, and the feature maps in dashed boxes (middle part) are weighted by output of spatial-temporal attention module. Best viewed on color display.}
  \label{FlowTrack_training}
\end{figure*}

The overall training framework of our tracker consists of FeatureNet (feature extraction sub-network), FlowNet, warping module, spatial-temporal attention module and CF tracking layer. As shown in Figure~\ref{FlowTrack_training}, overall training architecture adopts Siamese network consisting of historical and current branches. In historical branch, appearance features and flow information are extracted by the FeatureNet and FlowNet at first. Then previous frames at predefined intervals (5 frames in experiments, $T=6$) is warped to $t-1$ frame guided by flow information. Meanwhile, a spatial-temporal attention module is designed to weight the warped feature maps. In another branch, the feature maps of current frame is extracted by FeatureNet.
Finally,  both two branches are fed into subsequent correlation filters layer for training. All the modules are differentiable and trained end-to-end.

\subsection{Correlation filter layer}
Discriminative correlation filters (DCF) with deep convolutional features have shown favorable performance in recent benchmarks \cite{c27, c28, c32}. Nonetheless, the chosen CNN features are always pre-trained in different tasks and individual components in tracking systems are learned separately, thus the achieved tracking results may be suboptimal. Recently, CFNet \cite{cfnet} and DCFNet \cite{dcfnet} interpret the correlation filters as a  differentiable layer in Siamese framework, thus performing end-to-end representation learning.

In DCF tracking framework, the aim is to learn a series of convolution filters $\textbf f$ from training samples ${(\textbf x_k, \textbf y_k)}_{k=1:t}$. Each sample is extracted using the FeatureNet from an image region. Assuming sample has the spatial size $M \times N$, the output has the spatial size $m \times n$ ($m=M / stride_M, n=N / stride_N$). The desired output $\textbf y_k$ is a response map which includes a target score for each location in the sample $\textbf x_k$. The response of the filters on sample $\textbf x$ is given by

\begin{equation}
\label{eq1}
R(\textbf x) = \sum_{l=1}^d\varphi^l(\textbf x)*\textbf f^l
\end{equation}where $\varphi^l(\textbf x)$ and $\textbf f^l$ is $l$-th channel of extracted CNN features and desired filters, respectively, $*$ denotes circular correlation operation. The filters can be trained by minimizing error which is obtained between the response $R(\textbf x_k)$ on sample $\textbf x_k$ and the corresponding Gaussian label $\textbf y_k$:

\begin{equation}
\label{eq2}
e = \sum_k{||R(\textbf x_k) - \textbf y_k||}^2 + \lambda\sum_{l=1}^d{||\textbf f^l||}^2
\end{equation}The second term in~(\ref{eq2}) is a regularization with a weight parameter $\lambda$. The solution can be gained as \cite{c45}:

\begin{equation}
\label{eq3}
\textbf f^l =\mathcal{F}^{-1}\left(\frac{\hat{\varphi}^l(\textbf x) \odot \hat{\textbf y}^* }{\sum_{k=1}^D  \hat{\varphi}^k(\textbf x)\odot (\hat{\varphi}^k(\textbf x))^* + \lambda }\right)
\end{equation}where the hat symbol represents the discrete Fourier transform $\mathcal{F}$ of according variables, $^*$ represents the complex conjugate of according variables, and $\odot$ denotes Hadamard product.

In test stage, the trained filters are used to evaluate an image patch centered around the predicted target location:

\begin{equation}
\label{eq4}
R(\textbf z) = \sum_{l=1}^d\varphi^l(\textbf z)*\textbf f^l
\end{equation}where $\varphi(\textbf z)$ denote the feature maps extracted from tracked target position of last frame including context.

In order to unify the correlation filters in an end-to-end network, we formulate above solution as correlation filters layer. Given the feature maps of search patch $\varphi(\textbf z)$, the loss function is formulated as:

\begin{equation}
\begin{aligned}
\label{eq5}
L(\bm{\theta}) =& ||R(\bm{\theta})-\tilde R||^2+\gamma ||\bm{\theta}||^2 \\
s.t. \quad R(\bm{\theta}) =& \sum_{l=1}^d\varphi^l(\textbf z,\bm{\theta})*\textbf f^l \\
\textbf f^l =&\mathcal{F}^{-1}\left(\frac{\hat{\varphi}^l(\textbf x,\bm{\theta}) \odot \hat{\textbf y}^* }{\sum_{k=1}^D  \hat{\varphi}^k(\textbf x,\bm{\theta}) \odot (\hat{\varphi}^k(\textbf x,\bm{\theta}))^* + \lambda }\right)
\end{aligned}
\end{equation}where $\tilde R$ is desired response, and it is a gaussian distribution centered at the real target location. The back-propagation of loss with respect to $\varphi(\textbf x)$ and $\varphi(\textbf z)$ are formulated as \cite{dcfnet}:

\begin{equation}
\begin{aligned}
\label{eq6}
\frac{\partial L}{\partial \varphi^l(\textbf x)} &=\mathcal{F}^{-1}\left(\frac{\partial L}{\partial (\hat{\varphi}^l(\textbf x))^*}+\left(\frac{\partial L}{\partial (\hat{\varphi}^l(\textbf x))}\right)^*\right)\\
\frac{\partial L}{\partial \varphi^l(\textbf z)} &=\mathcal{F}^{-1}\left(\frac{\partial L}{\partial (\hat{\varphi}^l(\textbf z))^*}\right)
\end{aligned}
\end{equation}

Once the back-propagation is derived, the correlation filters can be formulated as a layer in network, which is called CF layer in next sections.

\subsection{Aggregation using optical flow}
Optical flow encodes correspondences between two input images. We warp the feature maps from the neighbor frames to specified frame according to the flow:{\setlength\abovedisplayskip{5pt}
\setlength\belowdisplayskip{5pt}
\begin{equation}
\label{eq7}
\varphi_{i \rightarrow t-1} = \mathcal{W}(\varphi_i, Flow(I_i, I_{t-1}))
\end{equation}where $\varphi_{i\rightarrow t-1}$ denotes the feature maps warped from previous frame $i$ to specified $t-1$ frame. $Flow(I_i, I_{t-1})$ is the flow field estimated through a flow network \cite{flownet}, which projects a location $\textbf p$ in frame $i$ to the location $\textbf p+ \delta \textbf p$ in specified frame $t-1$. The warping operation is implemented by the bilinear function applied on all the locations for each channel in the feature maps. The warping in certain channel is performed as:

\begin{equation}
\label{eq8}
\varphi_{i \rightarrow t-1}^m(\textbf p) = \sum_\textbf q K(\textbf q, \textbf p+\delta \textbf p) \varphi_i^m(\textbf q)
\end{equation}where $\textbf p = (p_x,p_y)$ means 2D locations, and $\delta \textbf p = Flow(I_i, I_{t-1})(\textbf p)$ represents flow in according positions, $m$ indicates a channel in the feature maps $\varphi(\textbf x)$, $\textbf q = (q_x,q_y)$ enumerates all spatial locations in the feature maps, and $K$ indicates the bilinear interpolation kernel.

Since we adopt end-to-end training, the back-propagation of $\varphi_{i \rightarrow t-1}$ with respect to $\varphi_i$ and flow $\delta \textbf p$ (\ie $Flow(I_i, I_{t-1})(\textbf p)$) is derived as:

\begin{equation}
\centering
\label{eq9}
\begin{aligned}
\frac{\partial \varphi_{i \rightarrow t-1}^m(\textbf p)}{\partial \varphi_i^m(\textbf q)} =  &K(\textbf q, \textbf p+\delta \textbf p)\\
\frac{\partial \varphi_{i \rightarrow t-1}^m(\textbf p)}{\partial Flow(I_i, I_{t-1})(\textbf p)} = &\sum_\textbf q    \frac   {\partial K(\textbf q, \textbf p+\delta \textbf p)} { \partial \delta \textbf p} \varphi_i^m(\textbf q)
\end{aligned}
\end{equation}

Once the feature maps in previous frames are warped to specified frame, they provide diverse information for same object instance, such as different viewpoints, deformation and varied illuminations. So appearance feature for tracked object can be enhanced by aggregating these feature maps. The aggregation results at are obtained as:

\begin{equation}
\label{eq10}
\varphi(\textbf x) = \overline{\varphi}_{t-1} = \sum_{i=t-T}^{t-1}  w_{i\rightarrow t-1} \varphi_{i\rightarrow t-1}
\end{equation}where $T$ is predefined intervals, $w_{i\rightarrow t-1}$ is adaptive weights at different spatial locations and feature channels. The adaptive weights are decided by proposed novel spatial-temporal attention mechanism which is described in detail in next subsection.

\subsection{Spatial-temporal attention}
The adaptive weights indicate the importance of aggregated frames at each spatial location and temporal channels. For spatial location, we adopt cosine similarity metric to measure the similarity between the warped features and the features extracted from the specified $t-1$ frame. For different channels, we further introduce temporal attention to adaptively re-calibrate temporal channels.
\subsubsection{Spatial attention}
Spatial attention indicates the different weights at different spatial locations. At first, a bottleneck sub-network projects the $\varphi$ into a new embedding $\varphi^e$, then the cosine similarity metric is adopted to measure the similarity between the warped features and the features extracted from the specified $t-1$ frame:

\begin{equation}
\label{eq11}
w_{i\rightarrow t-1} (\textbf p)= SoftMax \left(\frac{\varphi_{i\rightarrow t-1}^e(\textbf p) \varphi_{t-1}^e(\textbf p) }{\left| \varphi_{i\rightarrow t-1}^e(\textbf p)\right| \left|\varphi_{t-1}^e(\textbf p)\right| } \right)
\end{equation}where $SoftMax$ operation is applied at channels to normalize the weight $w_{i\rightarrow t-1}$ for each spatial location $\textbf p$ over the nearby frames. Intuitively speaking, in spatial attention, if the warped features $\varphi_{i\rightarrow t-1}^e(\textbf p)$ is close to the features $\varphi_{t-1}^e(\textbf p)$, it is assigned with a larger weight. Otherwise, a smaller weight is assigned.

\subsubsection{Temporal attention}

The weight $w_{i\rightarrow t-1}$ obtained by spatial attention has largest value at each position in $t-1$ frame because $t-1$ frame is most similar with its own according to cosine measurement. We further propose temporal attention mechanism to solve this problem by adaptively re-calibrating temporal channel as shown in Figure~\ref{FlowTrack_temporal_attention}. The channel number of spatial attention out is equal to the aggregated frame numbers $T$, and we expect to re-weight the channel importance by introducing temporal information.

Specifically, the output of spatial attention module is first passed through a global pooling layer to produce a channel-wise descriptor. Then three fully connected (FC) layers are added, in which learned for each channel by a self-gating mechanism based on channel dependence. This is followed by re-weighting the original feature maps to generate the output of temporal attention module. It is noting that our temporal attention mechanism is similar to SENet architecture in \cite{senet}, while more parameters is adopted in FC layers.

The weights in temporal frames (channels) are visualized to illustrate the results of our temporal attention. In Figure ~\ref{temporal_attention_weight}, the first and second row indicates the normal and challenging scenarios, respectively. As shown in top left corner in each frames, the weights are approximately equal in normal scenarios. In challenging scenarios, the weights are smaller in low quality frames while larger in high quality frames, which shows re-calibration role of the temporal attention module.

\begin{figure}[!tp]
  \centering
  \includegraphics[width=1\linewidth]{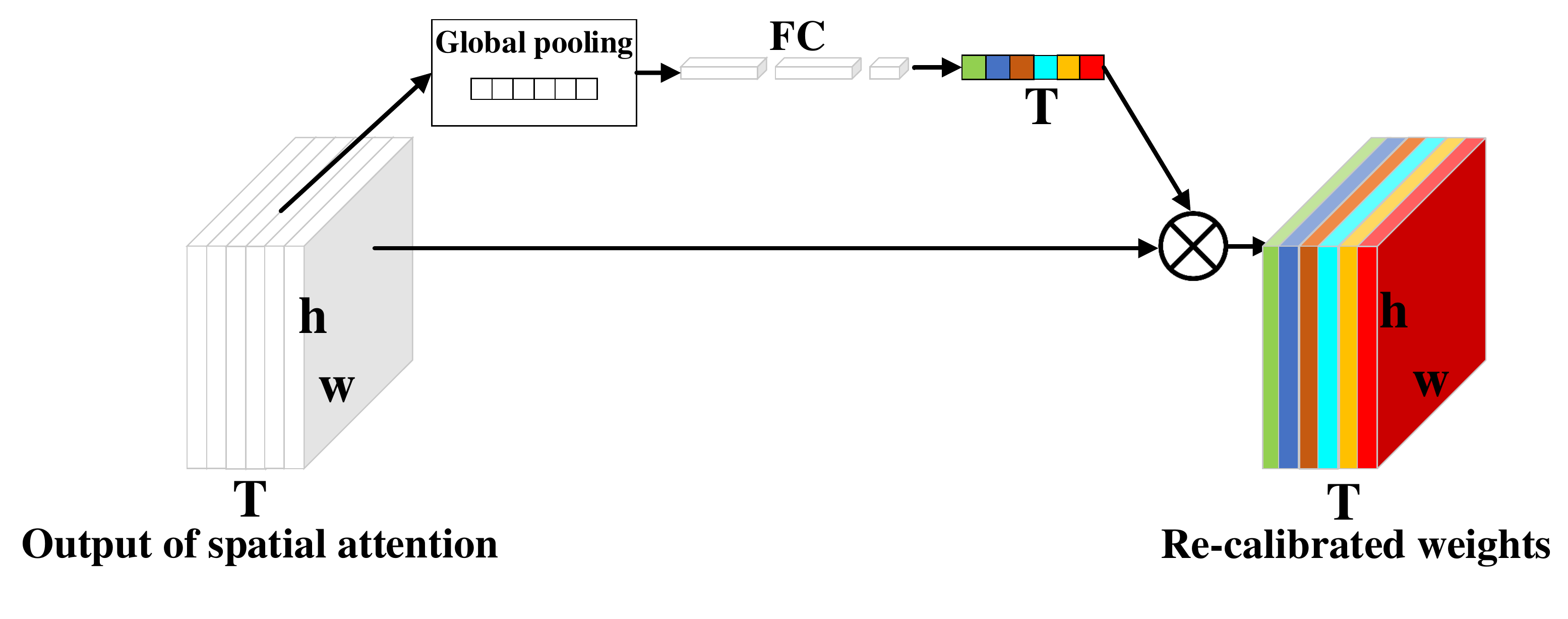}
  \caption{The temporal attention sub-network architecture. Channels with different colors are re-calibrated by different weights. Best viewed on color display.}
  \label{FlowTrack_temporal_attention}
\end{figure}

\begin{figure}[!tp]
  \centering
  \includegraphics[width=0.8\linewidth]{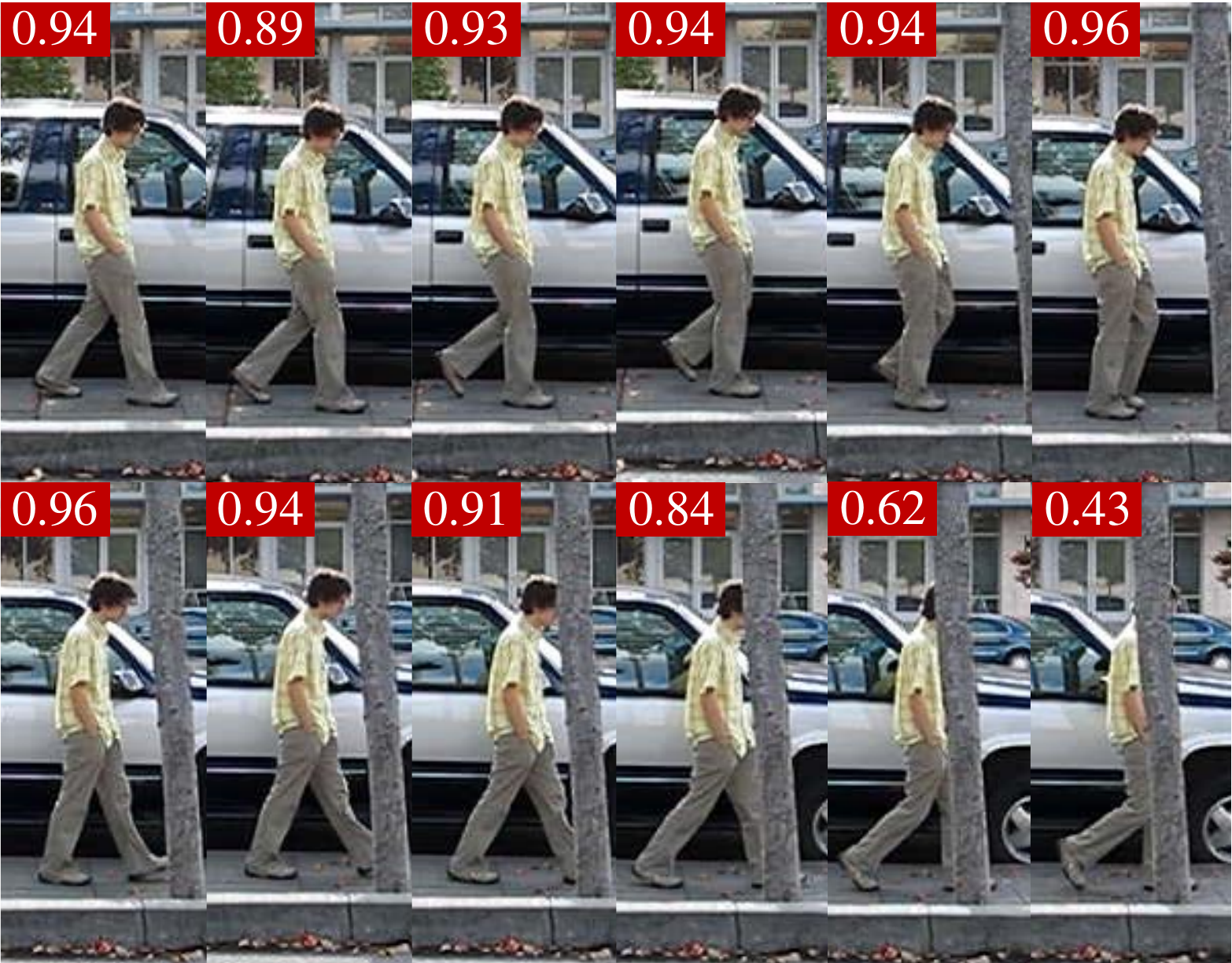}
  \caption{The visualization of weights in temporal frames (channels). The first and second row show normal and challenging scenarios, respectively. The number in top left corner indicates learned temporal weights. Best viewed on color display.}
  \label{temporal_attention_weight}
\end{figure}

\subsection{Online Tracking}
 In this subsection, tracking network architecture is described at first which is denoted as FlowTrack. Then we present the tracking process through the aspects of scale handing and model updating.

\paragraph{Tracking network architecture}
After off-line training as described above, the learned network is used to perform online tracking by equation~(\ref{eq4}). At first, the images are passed through trained FeatureNet and FlowNet. Then the feature maps in previous frames are warped to the current one according to flow information. Warped feature maps as well as the current frame's are embedded and then weighted using spatial-temporal attention. The estimation of the current target state is obtained by finding the maximum response in the score map. The CF layer in Figure~\ref{FlowTrack_training} is replaced by standard CF tracking module.


\paragraph {Model updating} Most of tracking approaches update their model in each frame or at a fixed interval \cite{c15, c25, c27, c30, eco}. However, this strategy may introduce false background information when the tracking is inaccurate, target is occluded or out of view.  In this paper, model updating is performed when criterions peak-versus-noise ratio (\emph{PNR}) and maximum value of response map are satisfied at the same time. Readers are referred to \cite{uct} for details. Only standard CF tracking module is updated as:

\begin{equation}
\label{eq12}
\textbf f^l =\mathcal{F}^{-1}\left(\frac{ \sum_{t=1}^p  \alpha_t \hat{\varphi}^l(\textbf x_t)\odot   \hat{\textbf y_t}^* }{\sum_{t=1}^p  \alpha_t (\sum_{k=1}^D  \hat{\varphi}^k(\textbf x_t)\odot (\hat{\varphi}^k(\textbf x_t))^* + \lambda) }\right)
\end{equation}where $\alpha_t$ represents the impact of sample $\textbf x_t$.

\paragraph {Scales} To handle the scale change, we follow the approach in \cite{scale} and use patch pyramid with the scale factors $\{a^s\left| \right. s=\lfloor -\frac{S-1}{2}\rfloor, \lfloor -\frac{S-3}{2}\rfloor,..., \lfloor \frac{S-1}{2}\rfloor\}$.

\section{Experiments}
Experiments are performed on four challenging tracking datasets: OTB2013 with 50 videos, OTB2015 with 100 videos, VOT2015 and VOT2016 with 60 videos. All the tracking results use the reported results to ensure a fair comparison.

\subsection{Implementation details}
We adopt three convolution layers ($3\times3\times128, 3\times3\times128, 3\times3\times96$) in FeatureNet, and FlowNet follows the implementation in \cite{flownet}. Embedding sub-network in spatial attention consists of three convolution layers ($1\times1\times64, 3\times3\times64, 1\times1\times256$) which are randomly initialized. Fully connected (FC) layers in temporal attention is set to $1\times1\times128, 1\times1\times128, 1\times1\times6$. First two and last FC layer are followed by ReLU and Sigmoid, respectively.  Our training data comes from VID \cite{vid}, containing the training and validation set. The frame number of aggregation is set to 5 ($T$ in Figure~\ref{FlowTrack_training} is set to 6). In each frame, patch is cropped around ground truth with a 1.56 padding and resized into $128*128$. We apply stochastic gradient descent (SGD) with momentum of 0.9 to end-to-end train the network and set the weight decay $\lambda$ to 0.005. The model is trained for 50 epochs with a learning rate of $10^{-5}$. In online tracking, scale step $a$ and number $S$ is set to 1.025 and 5, scale penalty and model updating rate is set to 0.9925 and 0.015.
The proposed FlowTrack is implemented using MatConvNet \cite{matconvnet} on a PC with an Intel i7 6700 CPU, 48 GB RAM, Nvidia GTX TITAN X GPU. Average speed of the tracker is 12 FPS and the code will be made publicly available.

\subsection{Results on OTB}
OTB2013 \cite{c14} contains 50 fully annotated sequences that are collected from commonly used tracking sequences. OTB2015 \cite{c9} is the extension of OTB2013 and contains 100 video sequences. Some new sequences are more difficult to track. The evaluation is based on two metrics: precision plot and success plot. The precision plot shows the percentage of frames that the tracking results are within certain distance determined by given threshold to the ground truth. The value when threshold is 20 pixels is always taken as the representative precision score. The success plot shows the ratios of successful frames when the threshold varies from 0 to 1, where a successful frame means its overlap is larger than this given threshold. The area under curve (AUC) of each success plot is used to rank the tracking algorithm.
\subsubsection{Results of OTB2013}
\label{resultsonOTB2013}
In this experiment, we compare our method against recent trackers that presented at top conferences and journals, including CREST (ICCV 2017) \cite{crest}, MCPF (CVPR 2017) \cite{mcpf}, UCT (ICCV 2017 Workshop) \cite{uct}, CACF (CVPR 2017) \cite{cacf}, CFNet (CVPR 2017) \cite{cfnet}, CSR-DCF (CVPR 2017) \cite{CSR-DCF}, CCOT (CVPR 2016) \cite{c30}, SiamFC (ECCV 2016) \cite{c33}, Staple (CVPR 2016) \cite{c26}, SCT (CVPR 2016) \cite{c34}, HDT (CVPR 2016) \cite{c28},  DLSSVM (CVPR 2016) \cite{c41}, SINT+ (CVPR 2016) \cite{sint}, FCNT (ICCV 2015) \cite{c36}, CNN-SVM (ICML 2015) \cite{c40},HCF (ICCV 2015) \cite{c27}, KCF (T-PAMI 2015) \cite{c15}. The one-pass evaluation (OPE) is employed to compare these trackers.

\begin{figure}[!tp]
 \centering
\begin{minipage}[c]{4.2cm}
\includegraphics[width=4.6cm]{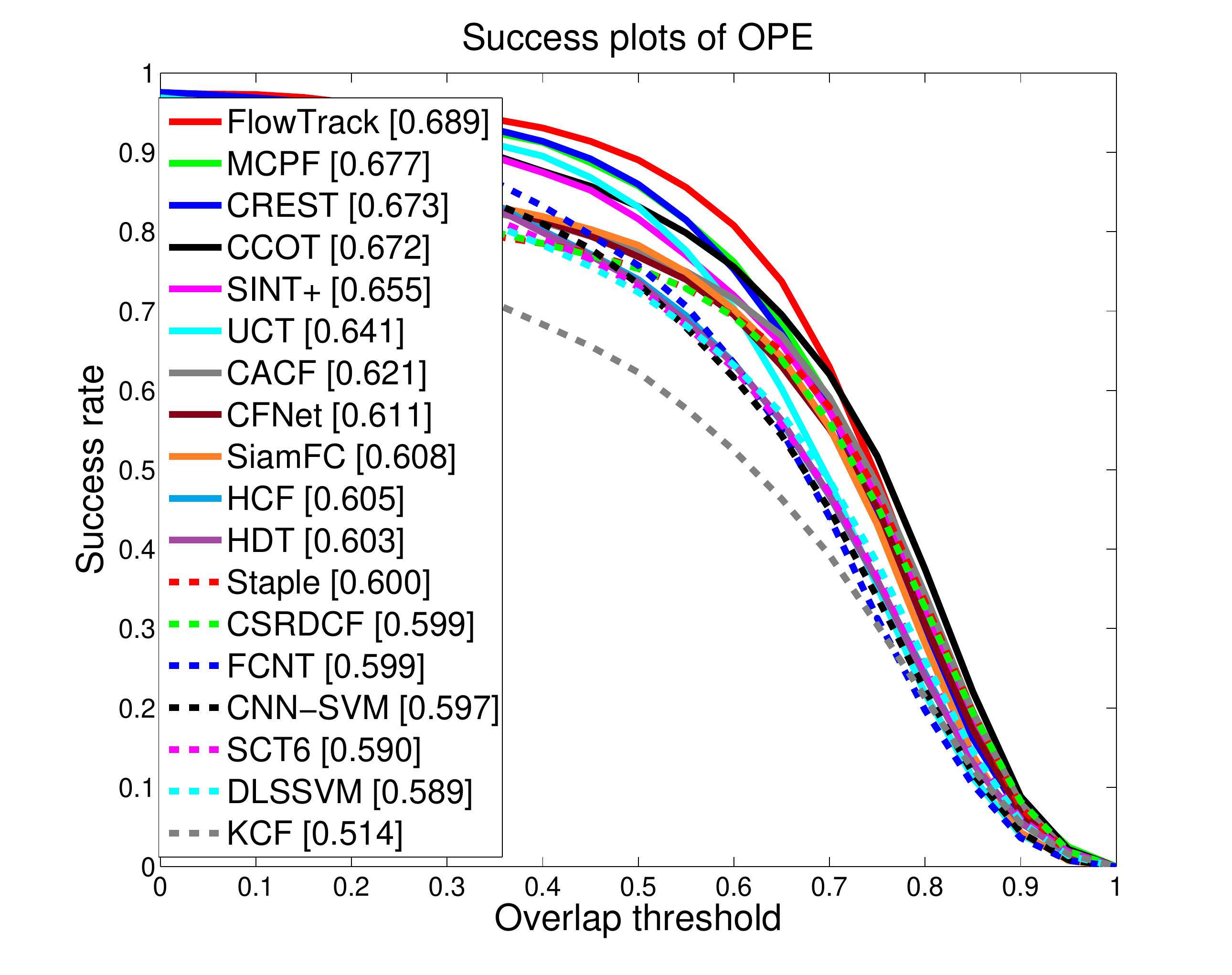}
\end{minipage}%
\begin{minipage}[c]{4.2cm}
\includegraphics[width=4.6cm]{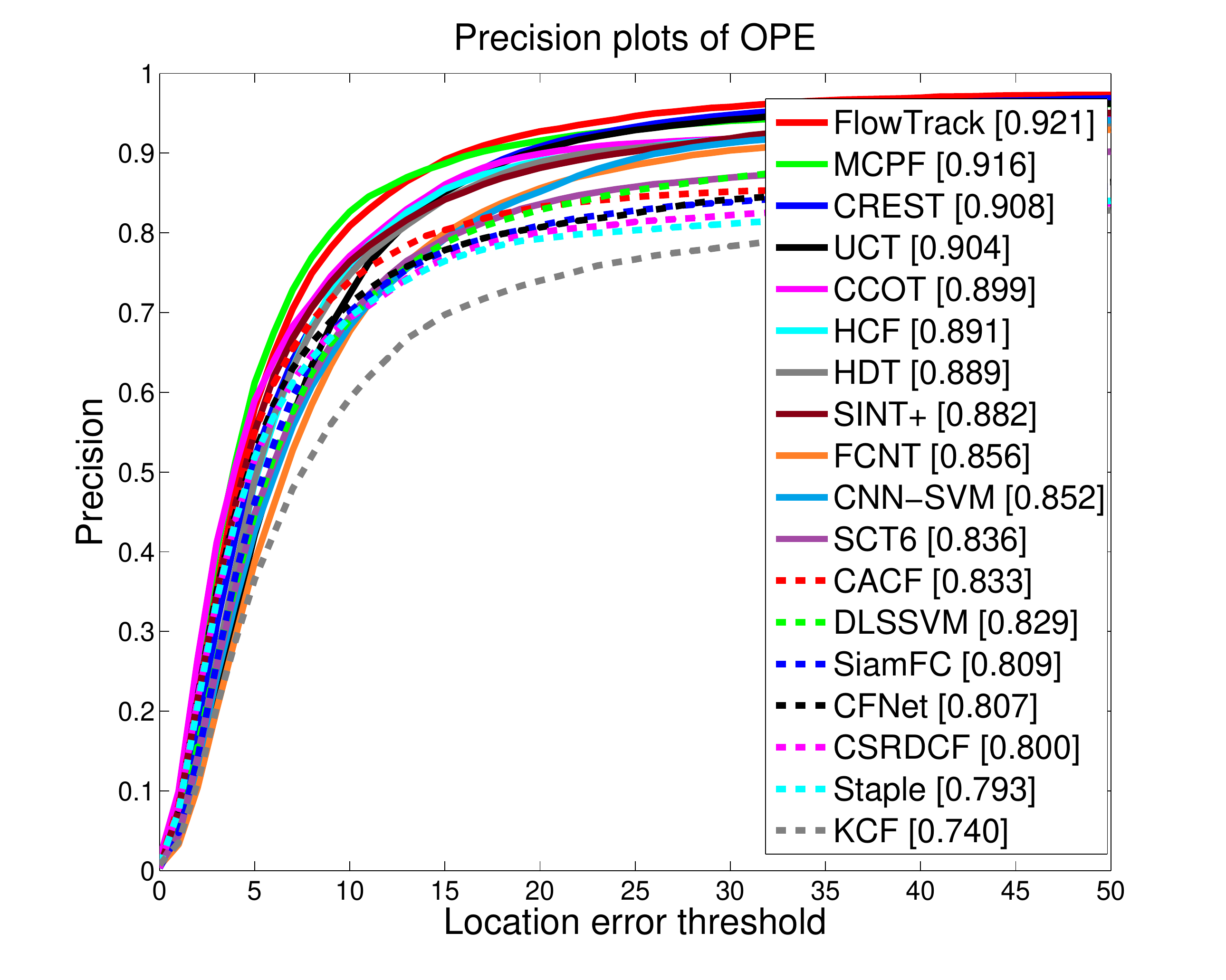}
\end{minipage}%
 \caption{Precision and success plots on OTB2013. The numbers in the legend indicate the representative precisions at 20 pixels for precision plots, and the area-under-curve scores for success plots. Best viewed on color display.}
 \label{OTB2013_OPE}
\end{figure}

 Figure~\ref{OTB2013_OPE} illustrates the precision and success plots based on center location error and bounding box overlap ratio, respectively. It clearly illustrates that our algorithm, denoted by FlowTrack, outperforms the state-of-the-art trackers significantly in both measures. In the success plot, our approach obtain an AUC score of 0.689, significantly outperforms the winner of VOT2016 (CCOT) and another tracker using flow information (SINT+). The improvement ranges are 1.7\% and 3.4\%, respectively. In the precision plot, our approach obtains a score of 0.921, outperforms CCOT and SINT+ by 2.2\% and 3.9\%, respectively.


The top performance can be attributed to that our method makes use of the rich flow information to improve the feature representation and the tracking accuracy. What is more, end-to-end training enables individual components in the tracking system are tightly coupled to work. By contrast, other trackers only consider appearance features, and hardly benefit from motion and inter-frame information. What is more, efficient updating and scale handling strategies ensure robustness of the tracker. It is worth noting that SINT+ adopts optical flow to filter out motion inconsistent candidates in Siamese tracking framework, while the optical flow is off-the-shelf and no end-to-end training is performed.

\subsubsection{Results of OTB2015}
 In this experiment, we compare our method against recent trackers, including CREST (ICCV 2017) \cite{crest}, CFNet (CVPR 2017) \cite{cfnet}, MCPF (CVPR 2017) \cite{mcpf}, UCT (ICCV 2017 Workshop) \cite{uct}, DSST (T-PAMI 2017) \cite{c45}, SiamFC (ECCV 2016) \cite{c33}, Staple (CVPR 2016) \cite{c26}, HDT (CVPR 2016) \cite{c28}, SINT (CVPR 2016) \cite{sint},  DLSSVM (CVPR 2016) \cite{c41}, CNN-SVM (ICML 2015) \cite{c40}, HCF (ICCV 2015) \cite{c27}, KCF (T-PAMI 2015) \cite{c15}. The one-pass evaluation (OPE) is employed to compare these trackers.

Figure~\ref{OTB2015_OPE} illustrates the precision and success plots of the compared trackers, respectively. The proposed FlowTrack approach outperforms all the other trackers in terms of success and precision scores. Specifically, our method achieves a success score of 0.655, which outperforms the MCPF (0.628) and CREST (0.623) method with a large margin. For detailed performance analysis, we also report the results on various challenge attributes in OTB2015, such as occlusion, illumination variation, background clutter, etc. Figure~\ref{OTB2015_OPE_attributes} demonstrates that our tracker effectively handles these challenging situations while other trackers obtain lower scores. Results comparisons of our approach with three state-of-the-art trackers in the challenging scenario is shown in Figure~\ref{tracking_results}.

\begin{figure}[!tp]
 \centering
\begin{minipage}[c]{4.2cm}
\includegraphics[width=4.6cm]{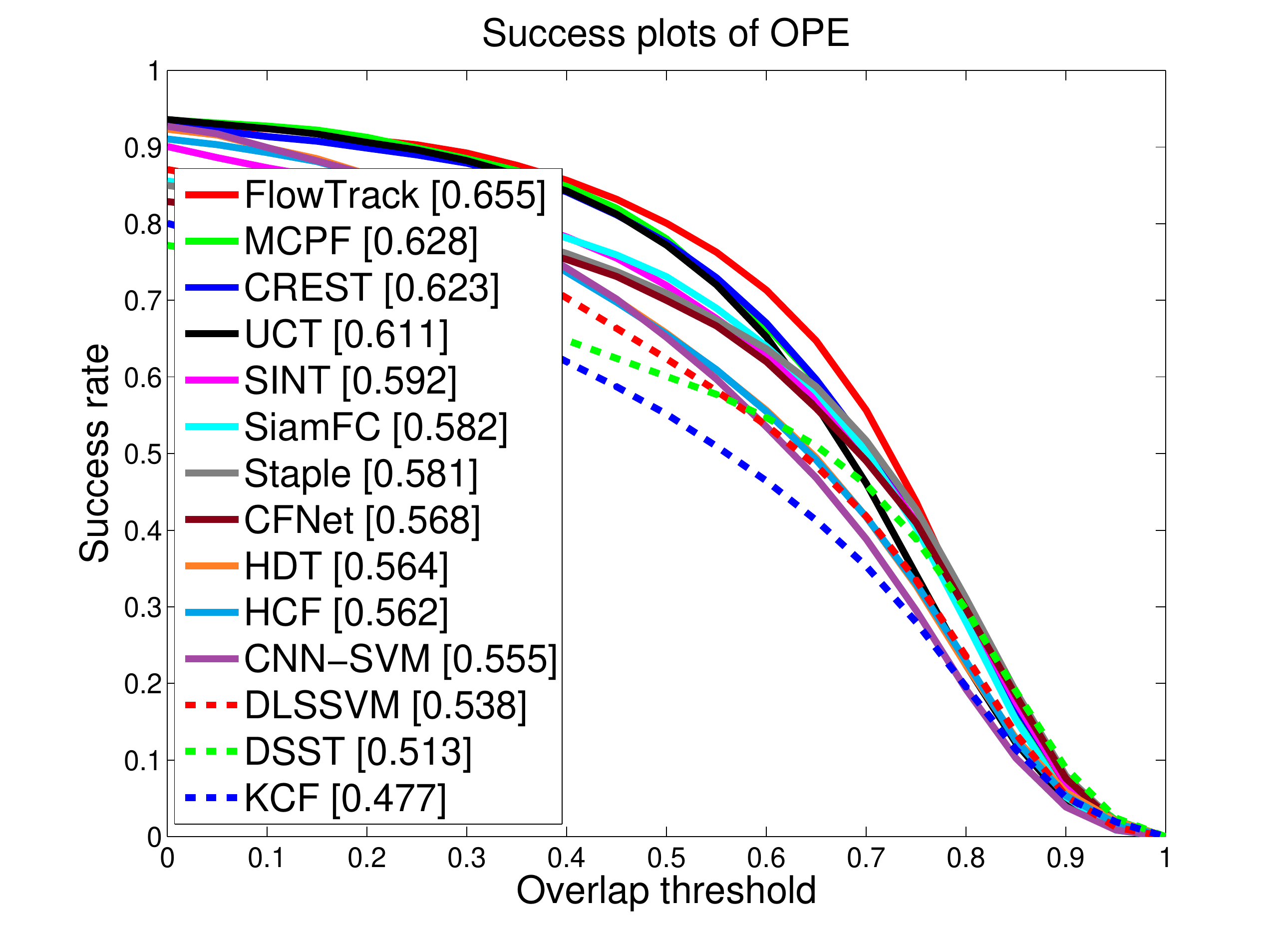}
\end{minipage}%
\begin{minipage}[c]{4.2cm}
\includegraphics[width=4.6cm]{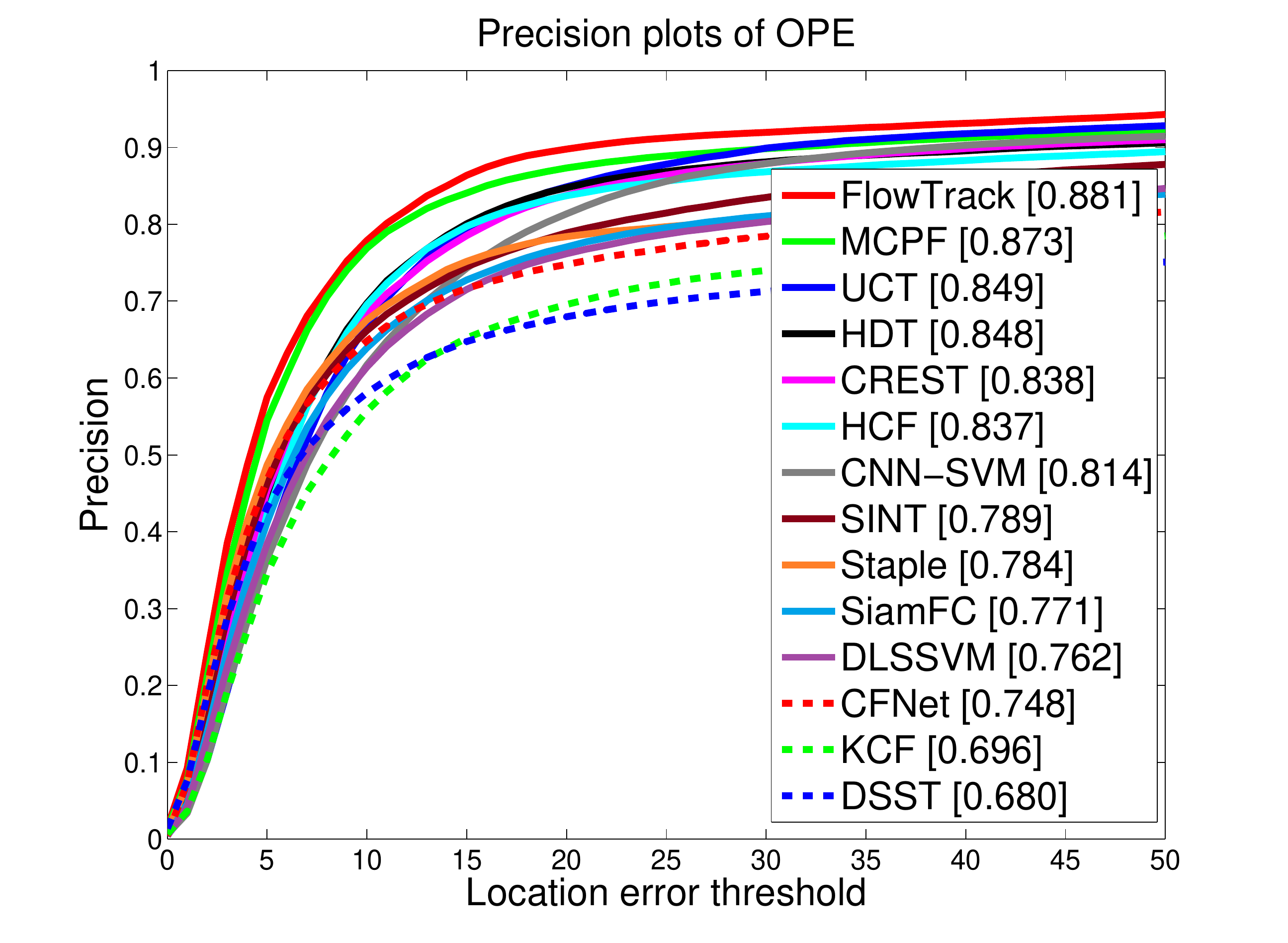}
\end{minipage}%
 \caption{Precision and success plots on OTB2015. The numbers in the legend indicate the representative precisions at 20 pixels for precision plots, and the area-under-curve scores for success plots. Best viewed on color display.}
 \captionsetup[subfloat]{captionskip=0pt,nearskip=0pt,farskip=0pt}
\captionsetup[figure]{position=bottom,belowskip=0pt,aboveskip=0pt}
\captionsetup[table]{belowskip=0pt,aboveskip=0pt}
\label{OTB2015_OPE}
\end{figure}

\begin{figure*}[!tp]
 \centering
\begin{minipage}[c]{3.53cm}
\includegraphics[width=3.85cm]{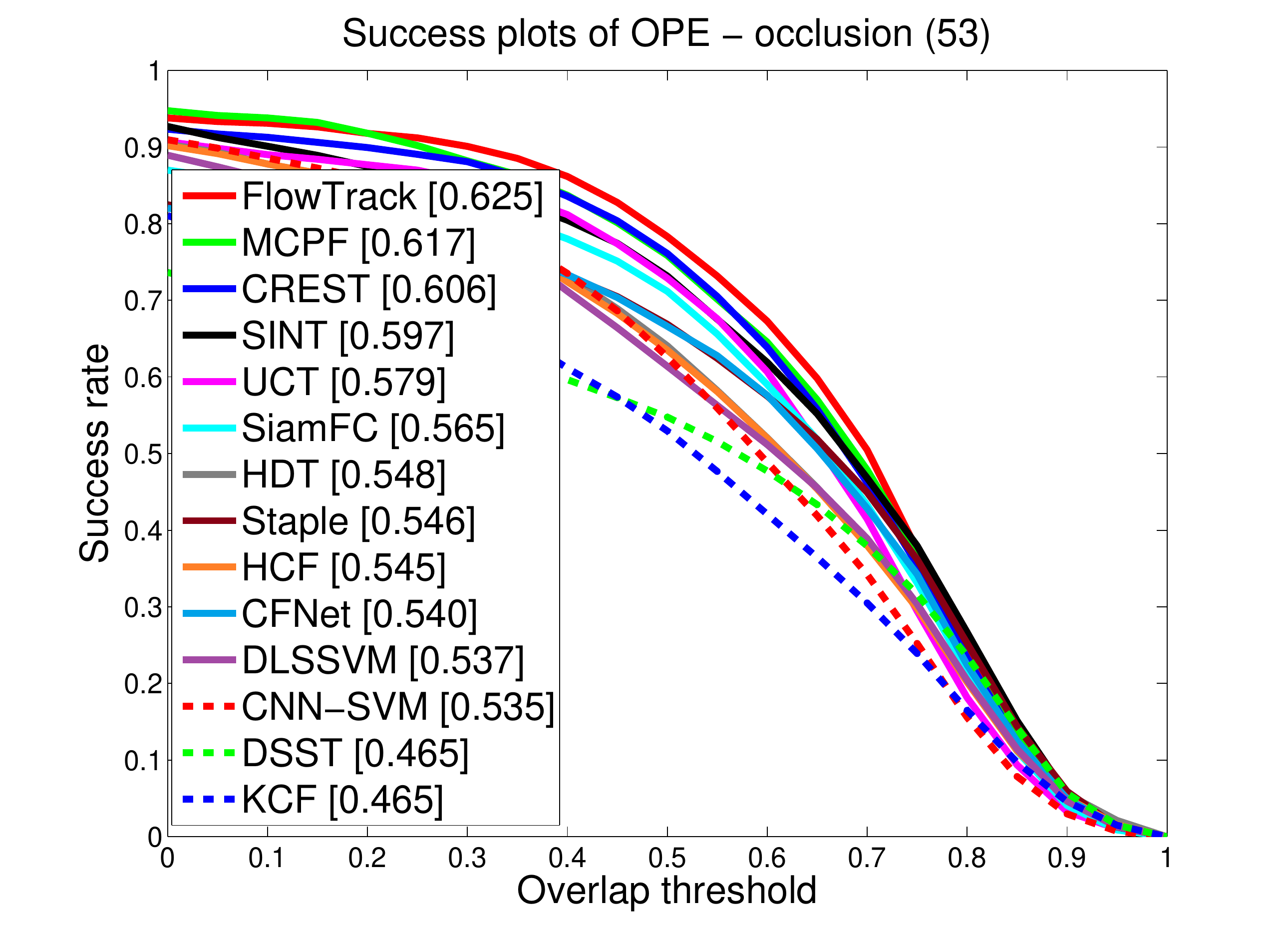}
\end{minipage}%
\begin{minipage}[c]{3.53cm}
\includegraphics[width=3.85cm]{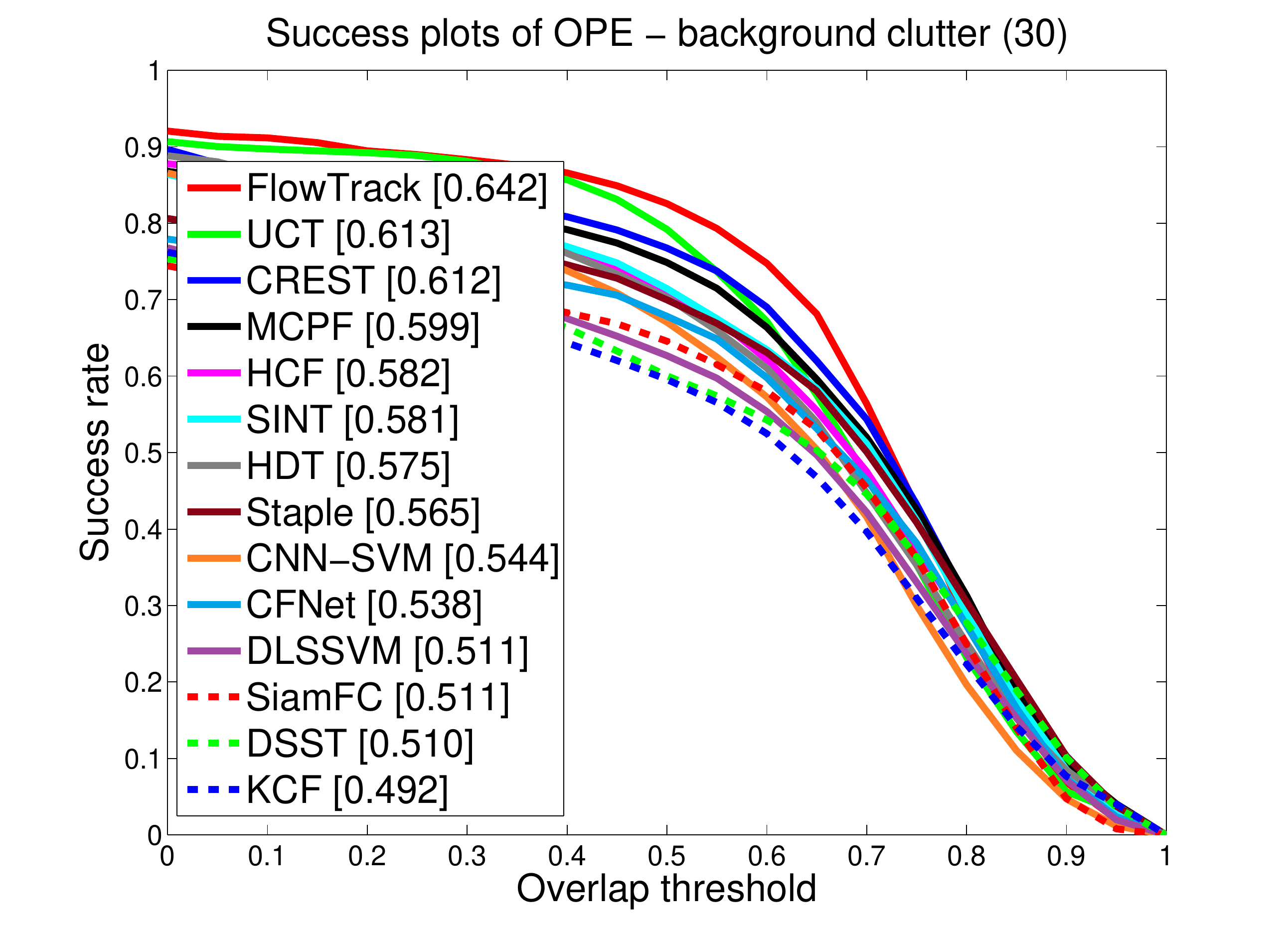}
\end{minipage}%
\begin{minipage}[c]{3.53cm}
\includegraphics[width=3.85cm]{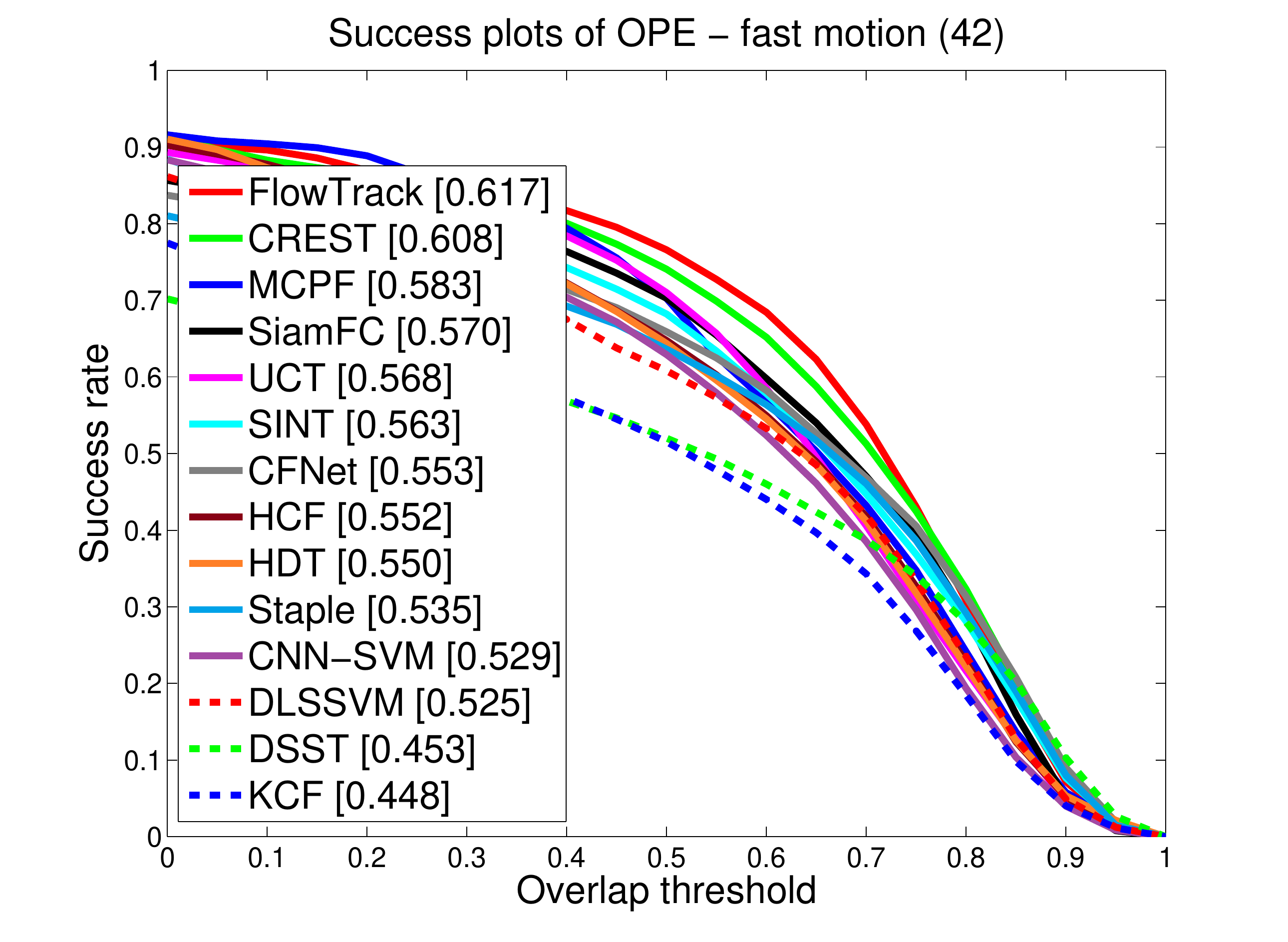}
\end{minipage}%
\begin{minipage}[c]{3.53cm}
\includegraphics[width=3.85cm]{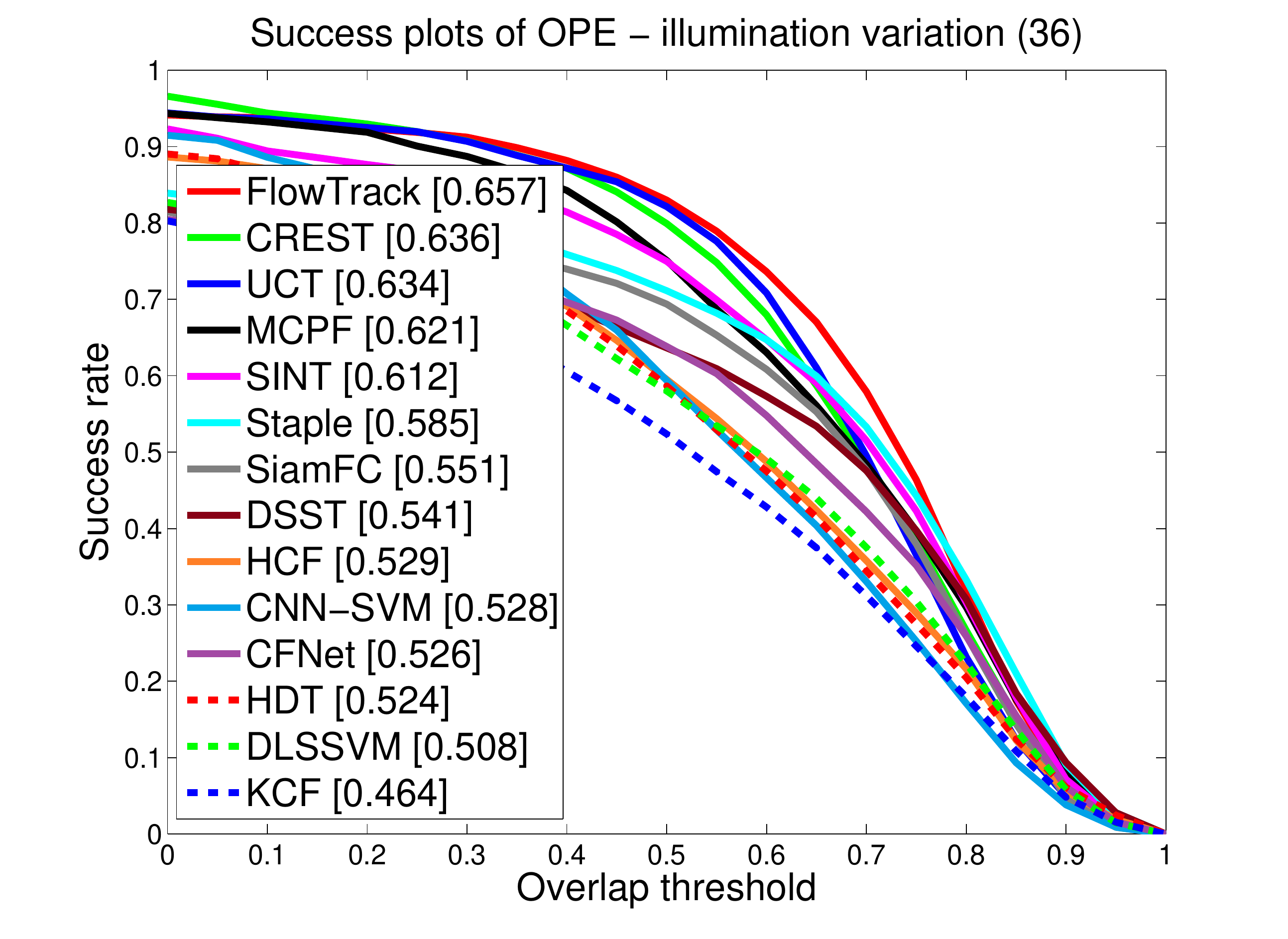}
\end{minipage}%
\begin{minipage}[c]{3.53cm}
\includegraphics[width=3.85cm]{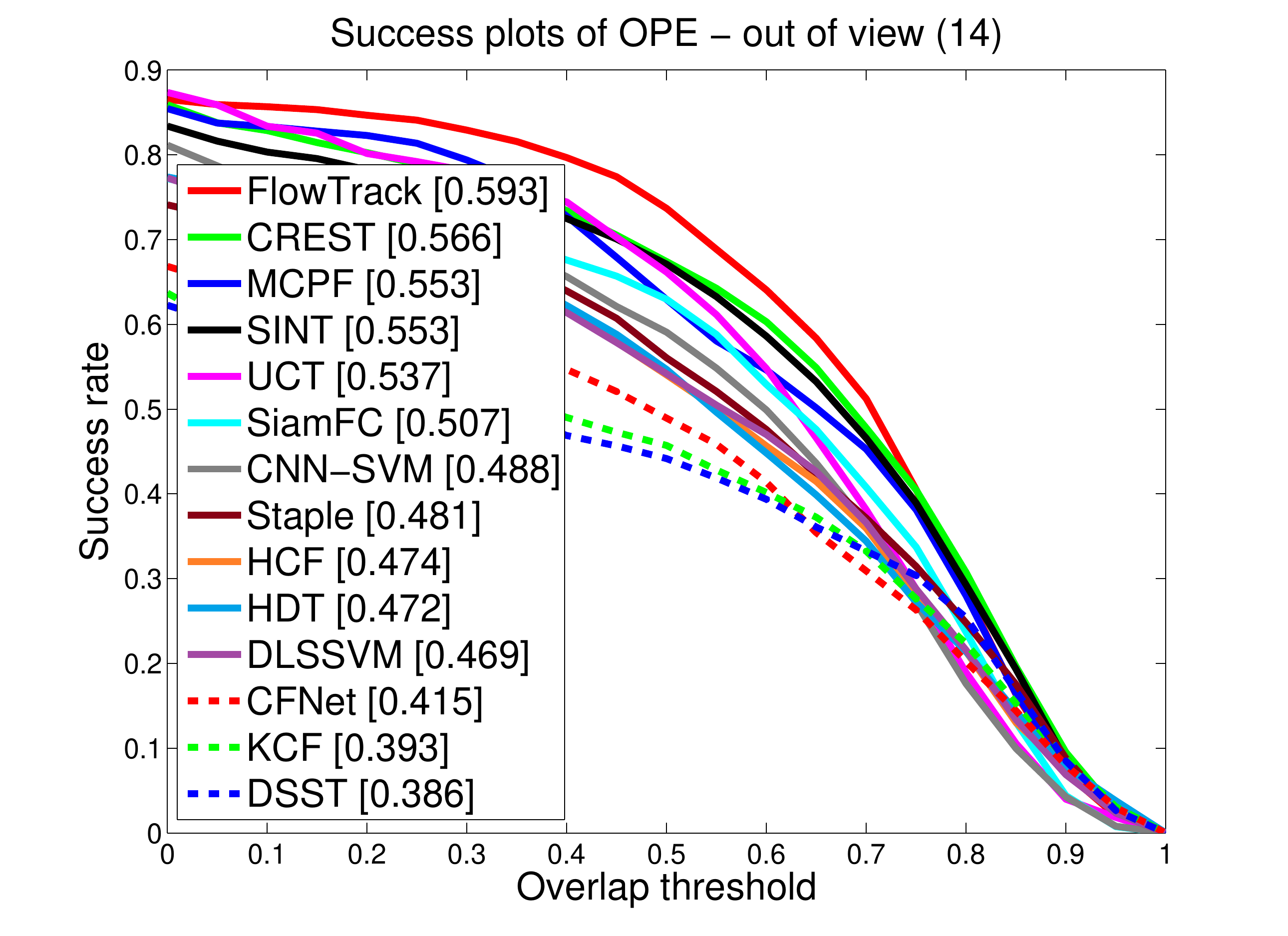}
\end{minipage}%
\caption{Success plots with attributes on OTB2015. Best viewed on color display.}
\setlength\abovecaptionskip{1pt}
\setlength\belowcaptionskip{-5pt}
\setlength{\intextsep}{0pt plus 1pt minus 1pt}
\setlength{\textfloatsep} {0pt plus 2pt minus 2pt}
\setlength\abovecaptionskip{0pt}
\setlength\belowcaptionskip{-5pt}
\label{OTB2015_OPE_attributes}
\end{figure*}

\subsection{Results on VOT}
The Visual Object Tracking (VOT) challenges are well-known competitions in tracking community, which have held several times from 2013 and their results will be reported at ICCV or ECCV. In this subsection, we compare our method, FlowTrack, with entries in VOT2015 \cite{c10} and VOT2016 \cite{c11}.

\subsubsection{Results of VOT2015}

\begin{figure}[tp]
 \centering
\begin{minipage}[c]{8cm}
\includegraphics[width=7.5cm]{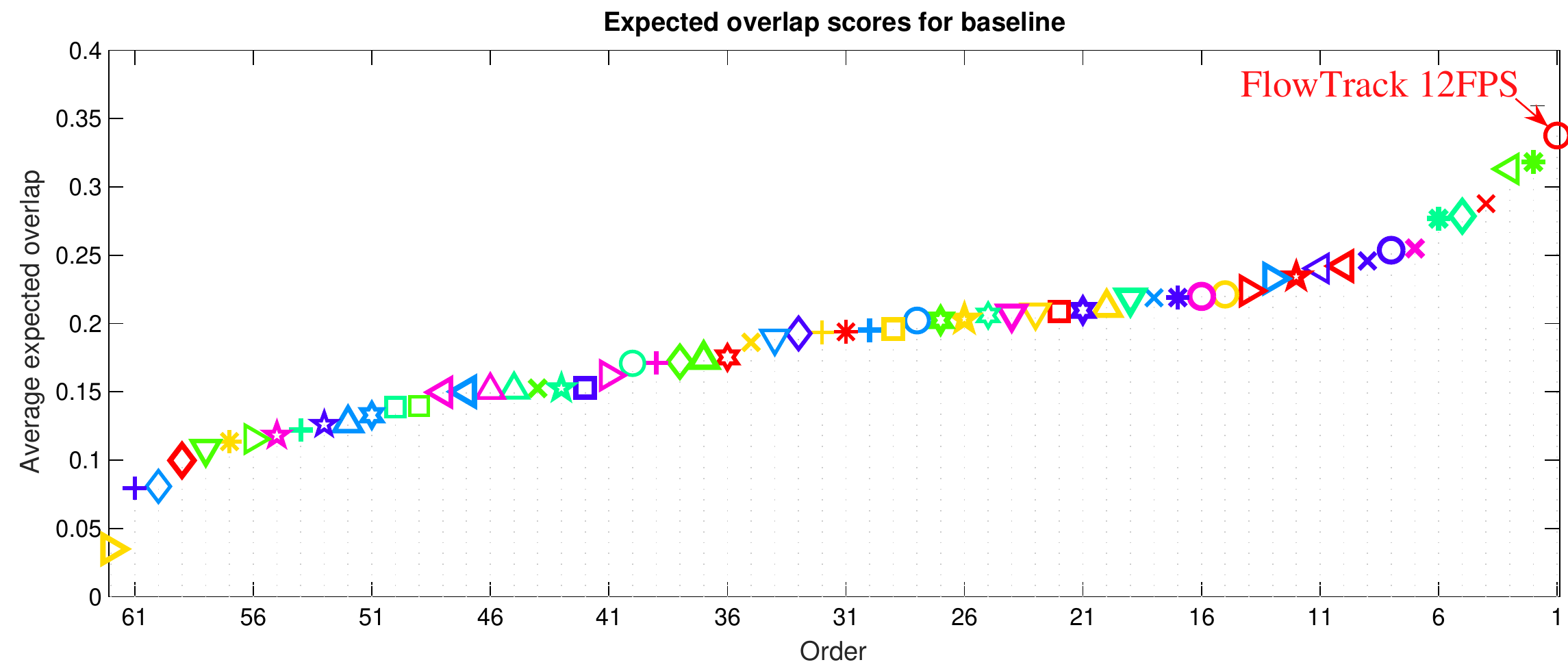}
\end{minipage}%

\begin{minipage}[c]{8cm}
\includegraphics[width=7.5cm]{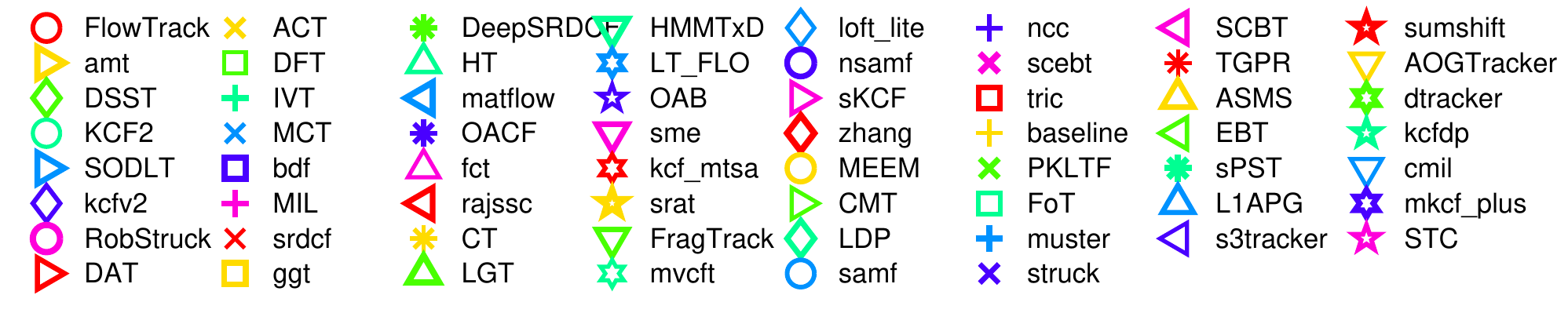}
\end{minipage}%
 \caption{EAO ranking with trackers in VOT2015. The better trackers are located at the right. Best viewed on color display.}
 \label{VOT2015_eao}
\end{figure}

\begin{table}[tp]
\scriptsize
  \centering
  \caption{ Comparisons with top trackers in VOT2015. {\color{red}Red}, {\color{green}green} and {\color{blue}blue} fonts indicate \emph{1st, 2nd, 3rd} performance, respectively. Best viewed on color display.}
\begin{tabular}{cccc}
\hline
\bf Trackers & \bf EAO & \bf Accuracy & \bf Failures  \\
\hline
\textbf{FlowTrack} & \color{red} 0.3405 & \color{red} 0.57 & \color{red} 0.95  \\\hline
\textbf{DeepSRDCF} & \color{green}0.3181 & \color{green}0.56 & \color{blue}1.05  \\\hline
\textbf{EBT}       & \color{blue}0.3130 & 0.47 & \color{green}1.02  \\\hline
\textbf{srdcf}     & 0.2877 & \color{green}0.56 & 1.24  \\\hline
\textbf{LDP}       & 0.2785 & 0.51 & 1.84  \\\hline
\textbf{sPST}      & 0.2767 & \color{blue}0.55 & 1.48   \\\hline
\textbf{scebt}     & 0.2548 & \color{blue}0.55 & 1.86  \\\hline
\textbf{nsamf}     & 0.2536 & 0.53 & 1.29  \\\hline
\textbf{struck}    & 0.2458 & 0.47 & 1.61  \\\hline
\textbf{rajssc}    & 0.2458 & \color{red}0.57 & 1.63   \\\hline
\textbf{s3tracker} & 0.2420 & 0.52 & 1.77 \\\hline
\end{tabular}
   \label{vot2015_table}
\end{table}

VOT2015 \cite{c10} consists of 60 challenging videos that are automatically selected from a 356 sequences pool. The trackers in VOT2015 is evaluated by expected average overlap (EAO) measure, which is the inner product of the empirically estimating the average overlap and the typical-sequence-length distribution. The EAO measures the expected no-reset overlap of a tracker run on a short-term sequence. Besides, accuracy (mean overlap) and robustness (average number of failures) are also reported.

In VOT2015 experiment, we present a state-of-the-art comparison with the participants in the challenge according to the latest VOT rules (see http://votchallenge.net). Figure~\ref{VOT2015_eao} illustrates that our FlowTrack can rank \emph{1st} in 61 trackers according to EAO criterion. It is worth noting that MDNet \cite{c29} is not compatible with the latest VOT rules because of OTB training data. In Table~\ref{vot2015_table}, we list the EAO, accuracy and failures of FlowTrack and top 10 entries in VOT2015. FlowTrack rank \emph{1st} according to all 3 criterions. The top performance can be attributed to the associating of flow information and end-to-end training framework.

\subsubsection{Results of VOT2016}
The datasets in VOT2016\cite{c11} are the same as VOT2015, but the ground truth has been re-annotated. VOT 2016 also adopts EAO, accuracy and robustness for evaluations.

In experiment, we compare our method with participants in challenges. Figure~\ref{VOT2016_eao} illustrates that our FlowTrack can rank \emph{1st} in 70 trackers according to EAO criterion. It is worth noting that our method can operate at 12 FPS, which is 40 times faster than CCOT (0.3 FPS).
For detailed performance analysis, we further list accuracy and robustness of representative trackers in VOT2016. As shown in Table~\ref{vot2016_table}, the accuracy and robustness of proposed FlowTrack can rank \emph{1st} and \emph{2nd}, respectively.
\begin{figure}[!tp]
\begin{minipage}[c]{8cm}
\includegraphics[width=8cm]{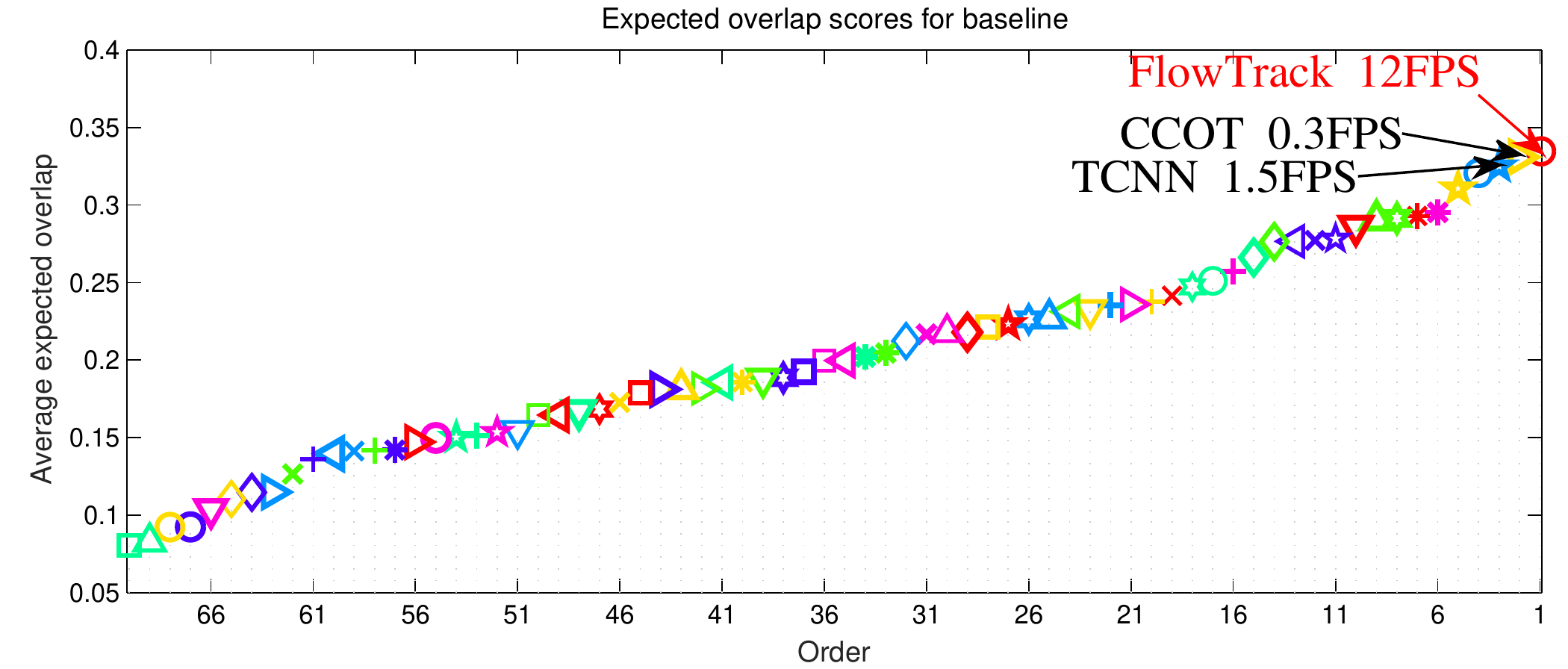}
\end{minipage}%

\begin{minipage}[c]{8cm}
\includegraphics[width=8cm]{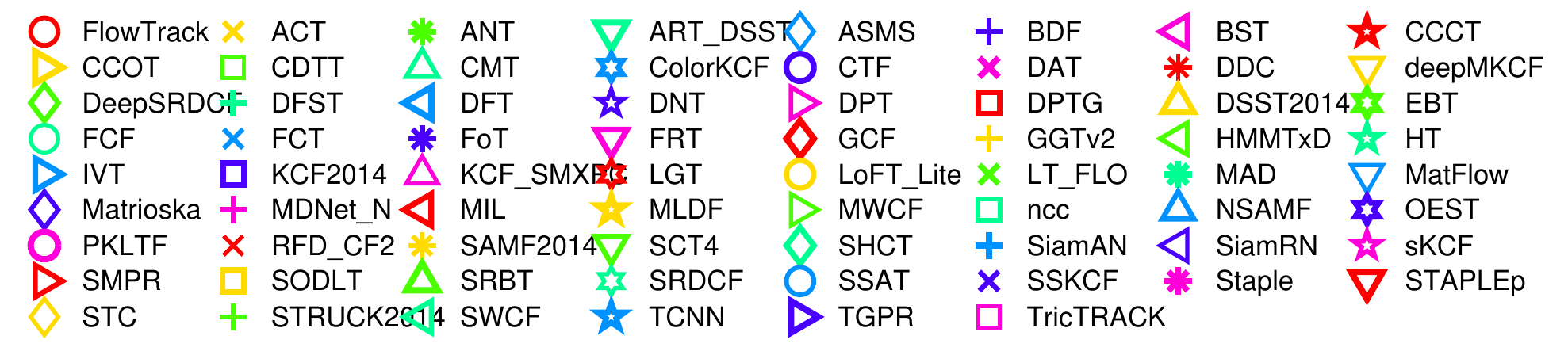}
\end{minipage}%
 \caption{EAO ranking with trackers in VOT2016. The better trackers are located at the right. Best viewed on color display.}
\captionsetup[figure]{position=bottom,belowskip=0pt,aboveskip=0pt}
\captionsetup[table]{belowskip=0pt,aboveskip=0pt}
 \label{VOT2016_eao}
\end{figure}

\begin{table}[!tp]
\scriptsize
  \centering
  \caption{ Comparisons with top trackers in VOT2016. {\color{red}Red}, {\color{green}green} and {\color{blue}blue} fonts indicate \emph{1st, 2nd, 3rd} performance, respectively. Best viewed on color display.}
\begin{tabular}{cccc}
\hline
\bf Trackers & \bf EAO & \bf Accuracy & \bf Robustness  \\
\hline
\textbf{FlowTrack}& \color{red}0.334   &\color{red}0.578     &\color{green}0.241\\\hline
\textbf{CCOT}     & \color{green}0.331     &0.539                &\color{red}0.238  \\\hline
\textbf{TCNN}     & \color{blue}0.325                &\color{green}0.554   &0.268 \\\hline
\textbf{Staple}   & 0.295    &0.544                 &0.378 \\\hline
\textbf{EBT}      & 0.291                &0.465                &\color{blue}0.252  \\\hline
\textbf{DNT}      & 0.278                &0.515                &0.329  \\\hline
\textbf{SiamFC}   & 0.277                &\color{blue}0.549   &0.382  \\\hline
\textbf{MDNet}    & 0.257                &0.541                &0.337  \\\hline
\end{tabular}
   \label{vot2016_table}
\end{table}

\subsection{Ablation analyses}

In this experiment, ablation analyses are performed to illustrate the effectiveness of proposed components. To verify the contributions of each component in our algorithm, we implement and evaluate four variations of our approach. At first, the baseline is implemented that no flow information is utilized(denoted by \emph{no flow}). Then the FlowNet is fixed to compare with end-to-end training (denoted by \emph{fix flow}).
To verify the superiority of proposed flow aggregation and spatial-temporal attention strategy, we fuse the warped feature maps by decaying with time (denoted by \emph{decay}). And the weight is obtained only by spatial attention, which is denoted as \emph{no\_ta} (means no temporal attention). Analyses results include OTB2013 \cite{c14}, OTB2015\cite{c9} VOT2015 \cite{c10}and VOT2016\cite{c11}. AUC means area under curve (AUC) of each success plot, and P20 represents precision score at 20 pixels.

As shown in Table~\ref{FlowTrack and its variations}, the performances of all the variations are not as good as our full algorithm (denoted by \emph{FlowTr}) and each component in our tracking algorithm is helpful to improve performance. Specifically, in terms of \emph{no flow} and \emph{FlowTr}, the associating and assembling of the flow information gains the performance with more than $6\%$ in all evaluation criterions. In terms of \emph{no flow}, \emph{fix flow} and \emph{FlowTr}, the performance of VOT even drops when FlowNet is added but fixed, which verifies the necessity of end-to-end training. Comparing \emph{decay} with \emph{FlowTr}, the superiority of proposed flow aggregation is verified by gaining the EAO in 2015 and 2016 by near 8\%. Besides, temporal attention further improves the tracking performance.

\begin{table}[!tp]
\scriptsize
  \centering
  \caption{ Performance on benchmarks of \emph{FlowTrack} and its variations}
 \begin{tabular}{C{0.72cm}C{0.75cm}C{0.75cm}C{0.75cm}C{0.75cm}C{0.75cm}C{0.75cm}}
    \hline
    &\bf \makecell[cc] {OTB2013 \\ AUC}  & \bf \makecell[cc] {OTB2013 \\ P20}  & \bf \makecell[cc] {OTB2015 \\ AUC}  & \bf \makecell[cc] {OTB2015 \\ P20} & \bf \makecell[cc] {VOT2015 \\ EAO} & \bf \makecell[cc] {VOT2016 \\EAO}\\
    \hline
    \emph{no flow} & 0.625 & 0.846 & 0.578 & 0.792 & 0.2637 & 0.2404 \\
    \emph{fix flow} & 0.617 & 0.853 & 0.583 & 0.813 & 0.2542 & 0.2291 \\
    \emph{decay}  & 0.637 & 0.868 & 0.586 & 0.793 & 0.2584 & 0.2516 \\
    \emph{no\_ta}   & 0.667 & 0.874 & 0.642 & 0.865 & 0.3109 & 0.2712 \\
    \emph{FlowTr}     & 0.689 & 0.921 & 0.655 & 0.881 & 0.3405 & 0.3342 \\
    \hline
  \end{tabular}

   \label{FlowTrack and its variations}
\end{table}

\subsection{Qualitative Results}
To visualize the superiority of flow correlation filters framework, we show examples of FlowTrack results compared to recent trackers on challenging sample videos.
As shown in Figure~\ref{tracking_results}, the target in sequence \emph{singer2} undergoes severe deformation. CCOT and CFNet lose the target from \emph{\#54} and CREST can not fit the scale change. In contrast, the proposed FlowTrack results in successful tracking in this sequence because feature representation is enhanced using flow information. \emph{skating1} is a sequences with attributes of illumination and pose variations, and proposed method can handle these challenges while CCOT drift to background. In sequence \emph{carscale}, only FlowTrack can handle the scale challenges in \emph{\#197} and \emph{\#252}. In background clutter of sequence \emph{bolt2}, FlowTrack tracks the target successfully while compared approaches drift to distracters.

\section{Conclusions}
In this work, we propose an end-to-end framework for tracking which makes use of the rich flow information in consecutive frames. Specifically, the frames in certain intervals are warped to specified frame using flow information and then they are aggregated for consequent correlation filter tracking. For adaptive aggregation, a novel spatial-temporal attention mechanism is developed. The effectiveness of our approach is validated in OTB and VOT datasets.

{\small
\bibliographystyle{ieee}

\end{document}